\documentclass[APA,LATO1COL]{WileyNJD-v2}
\graphicspath{ {./plots/} }

\articletype{Article Type}%

\received{26 April 2016}
\revised{6 June 2016}
\accepted{6 June 2016}

\usepackage{amsmath}
\usepackage{amssymb}
\usepackage{mathtools}
\usepackage{amsthm}
\usepackage{algcompatible}

\begin{document}
\title{The ART of Transfer Learning: An Adaptive and Robust Pipeline}

\author[1]{Boxiang Wang}

\author[2]{Yunan Wu*}

\author[3]{Chenglong Ye}

\authormark{Wang \textsc{et al}}

\address[1]{\orgdiv{Department of Statistics and Actuarial Science}, \orgname{University of Iowa}, \orgaddress{\state{IA}, \country{USA}}}

\address[2]{\orgdiv{Department of Mathematical Sciences}, \orgname{University of Texas at Dallas}, \orgaddress{\state{TX}, \country{USA}}}

\address[3]{\orgdiv{Dr. Bing Zhang Department of Statistics}, \orgname{University of Kentucky}, \orgaddress{\state{KY}, \country{USA}}}

\corres{*Yunan Wu. \email{yunan.wu@utdallas.edu}}

\abstract[Abstract]{Transfer learning is an essential tool for improving the performance of primary tasks by leveraging information from auxiliary data resources. In this work, we propose Adaptive Robust Transfer Learning (ART), a flexible pipeline of performing transfer learning with generic machine learning algorithms. We establish the non-asymptotic learning theory of ART, providing a provable theoretical guarantee for achieving adaptive transfer while preventing negative transfer. Additionally, we introduce an ART-integrated-aggregating machine that produces a single final model when multiple candidate algorithms are considered. We demonstrate the promising performance of ART through extensive empirical studies on regression, classification, and sparse learning. We further present a real-data analysis for a mortality study.}


\maketitle

\section{Introduction}
\label{Introduction}
The age of rapid technological change is unfolding in real time, empowering the collection of massive amounts of data in a variety of fields. Despite this, many fields still struggle with data acquisition with limited sample sizes, particularly in experiments that involve human or animal subjects and can be prohibitively expensive. To improve the performance of the primary task on those occasions, transfer learning has been widely advocated as a means of leveraging knowledge from available auxiliary data that are different while related to the primary data. Successful applications of transfer learning in data-scarce fields include drug development \citep{TWW17}, clinical trials \citep{BS19}, and material sciences \citep{HAG17}, among others. For an overview of transfer learning methodologies and applications, interested readers may refer to survey papers by \citet{PY09}, \citet{WKW16}, \citet{NLW20}, and \citet{ZQD20}.

Although transfer learning has achieved pervasive success and shows great promise, there is no guarantee that it will always improve performance -- there is no free lunch for transfer learning. When a large discrepancy exists between the primary and auxiliary data, the performance of the primary estimator is likely to be negatively affected by auxiliary data. This phenomenon is referred to as ``negative transfer" \citep{RMK05, WDP19}. Therefore, the success of a transfer learning method largely depends on its ability to be robust against negative transfer. As quoted from the survey paper \citet{ZQD20}, ``The negative transfer still needs further systematic analyses." 

In recent literature, the study of negative transfer has been embraced from the perspective of statistical guarantees for transfer learning. \citet{Bastani2021} studied estimation and prediction in high-dimensional linear models with one informative auxiliary data, where the sample size of the auxiliary data is required to be larger than its dimension. \citet{LCL21} proposed trans-lasso under a more general setting allowing for multiple auxiliary data, which can be even high-dimensional, i.e., the size can be smaller than the dimension. Trans-lasso has been shown to improve the learning efficiency with known informative auxiliary data and can be robust to non-informative auxiliary data. The idea of trans-lasso was further extended to generalized linear models in \citet{TF22}. \citet{cai2021transfer} proposed an estimation algorithm with a faster convergence rate than the minimax rate in the single study setting for high-dimensional Gaussian graphical models. For general function classes, \citet{tripuraneni2021provable} proposed meta-learning algorithms with theoretical underpinnings to estimate multiple linear regression models which share a common, low-dimensional linear representation. \citet{hanneke2022no} demonstrated that without access to distributional information, no algorithm could guarantee to improve the convergence rate with enlarging auxiliary data and primary data being fixed. 

The aforementioned theoretical studies mainly focused on certain methods under the framework of transfer learning. In this work, we consider a more general transfer learning framework called the Adaptive and Robust Transfer learning (ART) pipeline. ART is flexible and applicable to generic regression and classification methods rather than focusing on specific methods. The way that ART utilizes auxiliary data to supply information for the primary data is inspired by adaptive regression by mixing proposed in \citet{ARM}. With some random data splittings, primary and auxiliary data are aggregated through an exponential weighting scheme. We shall show a theoretical guarantee of robustness against the negative transfer for ART. The ART predictor has a prediction risk smaller than the best candidate's prediction risk plus a small penalty term, and the penalty term is the price to pay to achieve adaptivity without knowing which candidate is the best. In addition, when transfer learning is performed with multiple candidate algorithms, we propose a new method called ART-Integrated-Aggregating Machine (ART-I-AM) that aggregates those candidate algorithms under the framework of ART, automatically outputting a single model for the final prediction. Thus ARM-I-AM does not need the effort of selecting an algorithm, say by cross-validation, as is typically done in the standard practice. Further, when ART is applied for a sparse learning method that outputs sparse representations of the coefficient, e.g., lasso \cite{Tib96}, we present an ART variable importance measure that describes each predictor's contribution in the final predictions. 

The rest of this paper is organized as follows. Section~\ref{sec:methodology} presents the methodological details of ART. Section~\ref{sec:theory} establishes the theoretical properties of our proposed methods. Section~\ref{sec:simulations} contains simulation results,  and Section~\ref{sec:real_data} performs a real-data analysis for predicting the survival rate of ICU patients. Technical proofs and additional simulation results are presented in the supplemental file.

\section{Methodology}\label{sec:methodology}

\begin{algorithm}[t]
{\bfseries Input:} Primary data $T^{(0)}= \{(\mathbf{x}^{(0)}_i, y^{(0)}_i)\}_{i=1}^{n_0}$ and $M$ auxiliary data $T^{(m)}= \{(\mathbf{x}^{(m)}_i, y^{(m)}_i)\}_{i=1}^{n_m}$, for $m = 1, 2, \ldots, M$.
\begin{algorithmic}[1]
\STATE Split the primary data into two parts: $T^{(0)}_{\text{train}}= \{(x^{(0)}_i, y^{(0)}_i)\}_{i=1}^{{n_{0,\text{train}}}}$ and $T^{(0)}_{\text{test}}= \{(x^{(0)}_i, y^{(0)}_i)\}_{i={n_{0,\text{train}}}+1}^{n_0}$. 
\STATE Fit the model $\hat{g}^{(0)}$ by $\mathcal{A}(T^{(0)}_{\text{train}})$.
\FOR {$m=1$ {\bfseries to} $M$}
\STATE Stack $T^{(m)}$ with $T^{(0)}_{\text{train}}$ to have $\tilde{T}^{(m)}$, and obtain the model $\hat{g}^{(m)}$ by $\mathcal{A}(\tilde{T}^{(m)})$.
\STATE Set $w_{m, {n_{0,\text{train}}}+1}=\pi_{m}$ such that $\pi_{m} \geq 0$ and $\sum_{m=0}^{M} \pi_{m}=1$. 
\STATE For each $m \geq 0$ and ${n_{0,\text{train}}}+2 \leq i \leq n_0$, calculate the weights
\begin{equation}
\begin{aligned}
w_{m, i}=\frac{\pi_{m} \exp \left\{-\lambda \sum_{j=n_{0,\text{train}}+1}^{i-1} L\left(y^{(0)}_{j}, \hat{g}^{(m)}(\mathbf{x}_{j}^{(0)})\right)\right\}}{\sum_{m^{\prime}=0}^{M} \pi_{m^{\prime}} \exp \left\{-\lambda \sum_{j=n_{0,\text{train}}+1}^{i-1} L\left(y^{(0)}_{j}, \hat{g}^{(m')}(\mathbf{x}_{j}^{(0)})\right)\right\}}.
\end{aligned}
\label{eq:weights}
\end{equation}
\ENDFOR
\STATE Output the final model: 
\begin{equation}
\tilde{g}^{\text{ART}}(\mathbf{x})=\sum_{m=0}^{M}\left(\sum_{i=n_{0,\text{train}}+1}^{n_0} \frac{w_{m, i}}{n_0-{n_{0,\text{train}}}} \right) \hat{g}^{(m)}(\mathbf{x}).
\label{eq:final_wt}
\end{equation}
\end{algorithmic}
\caption{\textbf{ART}: \textbf{A}daptive and \textbf{R}obust \textbf{T}ransfer Learning Pipeline}
\label{alg:ART}
\end{algorithm}

Consider using a certain algorithm $\mathcal{A}$ to train a model $\hat{g}$ on \textit{primary data} $T^{(0)} = \{(\mathbf{x}^{(0)}_i, y^{(0)}_i)\}_{i = 1}^{n_0}$, where each $\mathbf{x}_i \in \mathbb{R}^p$ and $y_i \in \mathbb{R}$ for regression and $y_i \in \{0, 1\}$ for classification. The performance of $\hat{g}$ is assessed by its generalization error $EL(y, \hat{g}(\mathbf{x}))$ with a loss function $L$, where the expectation is taken over the same data generating distribution of the data in $T^{(0)}$. Despite the focus on binary labels for classification, the proposed method in this work can be naturally extended to multi-class classification. 

In this work, the ultimate goal is to enhance the performance of $\hat{g}$ with possible help from external data resources. Suppose $M$ \textit{auxiliary data}, namely, $T^{(m)} = \{(\mathbf{x}^{(m)}_i, y^{(m)}_i)\}_{i=1}^{n_m}$, $m = 1, 2, \ldots, M$, are available and they may potentially provide useful information for building $\hat{g}$. For simplicity, each $T^{(m)}$ is assumed to have the same predictors as $T^{(0)}$ and thus can be stacked upon $T^{(0)}$. 

A new pipeline called ART is introduced for transfer learning and presented as in Algorithm~\ref{alg:ART}. It is sketched as follows. We random split the primary data into two parts $T^{(0)}_{\text{train}}$ and $T^{(0)}_{\text{test}}$, and stack each of the auxiliary data with $T^{(0)}_{\text{train}}$ to have $\tilde{T}^{(m)}$. We fit each $\hat{g}^{(m)}$, $m = 0, 1, 2, \ldots, M$, by running the algorithm $\mathcal{A}$ on $T^{(0)}_{\text{train}}$ and each $\tilde{T}^{(m)}$, respectively. The models $\hat{g}^{(m)}$ are then aggregated according to an exponential-weighting scheme with the weights given in \eqref{eq:weights}, yielding the final ART estimate $\tilde{g}^{\text{ART}}$. For the sake of exposition, in Algorithm~\ref{alg:ART}, only a single random split on the primary data is presented, while the algorithm can be repeated to perform multiple random splits until the weights converge, making the output $\tilde{g}^{\text{ART}}$ more stable at the end.

The ART pipeline is generically applicable under both regression and classification settings, which differ in Algorithm~\ref{alg:ART} only in the choice of the loss function $L$ when assessing the accuracy and determining the weights. For regression problems, a common choice is $L(y, \hat{y}) = (y - \hat{y})^2$. For classification problems, with $y \in \{0, 1\}$, the cross entropy, $L(y, \hat{g}(\mathbf{x})) = -y \log \hat{g}(\mathbf{x}) -(1-y) \log (1-\hat{g}(\mathbf{x}))$, is employed, where $\hat{g}(\mathbf{x})$ is an estimate of the conditional probability $P(y = 1 | \mathbf{x})$ being yielded by $\mathcal{A}$. The conditional probabilities can be intrinsically estimated by some algorithms such as logistic regression, random forest, AdaBoost, etc; otherwise, we recommend using the \textit{classification calibration} approach as a post-hoc manner: for example, Platt scaling \citep{PO99}, which transfers the support vector machine \citep[SVM, ][]{CV95} outputs into probabilities. Other examples of classification calibration methods include beta calibration \citep{KSF17}, isotonic regression \citep{ZE02}, and confidence calibration \citep{GPS17}, among many others. 

\begin{remark}
Algorithm~\ref{alg:ART} can be simplified. The weights $w_{m,i}$ calculated in \eqref{eq:weights} are attributed to the cumulative performance of $\hat{g}^{(m)}$ on $\{(\mathbf{x}^{(0)}_t, y^{(0)}_t)\}_{t={n_{0,\text{train}}}+1}^{i}$ from $T^{(0)}_{\text{test}}$. To reduce the computational cost, we suggest the use of $\tilde{g}(\mathbf{x})=\sum_{m=0}^{M}w_m\hat{g}^{(m)}(\mathbf{x})$ and
\begin{equation*}
w_m = \frac{\pi_{m} e^{-\lambda \sum_{j={n_{0,\text{train}}}+1}^{n_0} L\left(y^{(0)}_{j}, \hat{g}^{(m)}(\mathbf{x}_{j}^{(0)})\right)}}{\sum_{m^{\prime}=0}^{M} \pi_{m^{\prime}} e^{-\lambda \sum_{j={n_{0,\text{train}}}+1}^{n_0} L\left(y^{(0)}_{j}, \hat{g}^{(m')}(\mathbf{x}_{j}^{(0)})\right)}},
\end{equation*}
in place of equations \eqref{eq:weights} and \eqref{eq:final_wt} in Algorithm~\ref{alg:ART}.
\end{remark}

\begin{remark}
Regarding the initial weighting choice $\pi_m$ for each auxiliary data, larger weights are typically recommended for more sizeable and trustful auxiliary data. Nevertheless, the equal weights $\pi_m\equiv 1/(M+1)$ 
usually work well in practice. As will be seen in Theorem \ref{thm:regression}, the upper bound of the generalization error of the final estimator $\tilde{g}^{\text{ART}}$ establishes a trade-off between the model complexity and accuracy. In addition, we recommend ${n_{0,\text{train}}}=\left\lfloor n_0/2\right\rfloor$, and shall discuss the choice of $\lambda$ in Remark~\ref{rem:lam}.
\end{remark}

Machine learning practitioners often need to choose from a wide variety of applicable algorithms to solve a certain problem. Given the flexibility of ART, we propose an ART-integrated-aggregating machine (ART-I-AM) in the context of transfer learning. ART-I-AM integrates multiple algorithms, e.g., the SVM, random forest, and boosting, in the ART pipeline, automatically producing a single output without extra tuning efforts. More details about ART-I-AM are summarized in Algorithm~2.


\begin{algorithm}[t]
{\bfseries Input:} Primary data $T^{(0)}= \{(\mathbf{x}^{(0)}_i, y^{(0)}_i)\}_{i=1}^{n_0}$ and $M$ auxiliary data $T^{(m)}= \{(\mathbf{x}^{(m)}_i, y^{(m)}_i)\}_{i=1}^{n_m}$, for $m = 1, 2, \ldots, M$. \\
Candidate algorithms: $\mathcal{A}^{1}, \mathcal{A}^{2}, \ldots, \mathcal{A}^{R}$. \\
\vspace{-0.5cm}
\begin{algorithmic}[1]
\STATE Random split the primary data into two parts: $T^{(0)}_{\text{train}}= \{(\mathbf{x}^{(0)}_i, y^{(0)}_i)\}_{i=1}^{{n_{0,\text{train}}}}$ and $T^{(0)}_{\text{test}}= \{(\mathbf{x}^{(0)}_i, y^{(0)}_i)\}_{i={n_{0,\text{train}}}+1}^{n_0}$. 
\FOR {$r=1$ {\bfseries to} $R$}
\STATE Fit the model $\hat{g}^{(0, r)}$ by $\mathcal{A}^{r}(T^{(0)}_{\text{train}})$.
\ENDFOR
\FOR {$m=1$ {\bfseries to} $M$}
\STATE Stack $T^{(m)}$ with $T^{(0)}_{\text{train}}$ to have $\tilde{T}^{(m)}$
\FOR {$r=1$ {\bfseries to} $R$}
\STATE Otain the model $\hat{g}^{(m, r)}$ by $\mathcal{A}^{r}(\tilde{T}^{(m)})$.
\STATE Set $w_{m, r, {n_{0,\text{train}}}+1}=\pi_{m, r}$ such that $\pi_{m, r} \geq 0$ and $\sum_{m=0}^{M}\sum_{r=1}^{R} \pi_{m, r}=1$. 
\STATE For each ${n_{0,\text{train}}}+2 \leq i \leq n_0$, calculate the weights
\begin{equation*}
\begin{aligned}
w_{m, r, i}=\frac{\pi_{m, r} \exp \left\{-\lambda \sum_{j=n_{0,\text{train}}+1}^{i-1} L\left(y^{(0)}_{j}, \hat{g}^{(m, r)}(\mathbf{x}_{j}^{(0)})\right)\right\}}{\sum_{m^{\prime}=0}^{M} \sum_{r^{\prime}=1}^{R}\pi_{m^{\prime}r^{\prime}} \exp \left\{-\lambda \sum_{j=n_{0,\text{train}}+1}^{i-1} L\left(y^{(0)}_{j}, \hat{g}^{(m', r')}(\mathbf{x}_{j}^{(0)})\right)\right\}}.
\end{aligned}
\end{equation*}
\ENDFOR
\ENDFOR
\STATE Output the final model: 
\begin{equation*}
\tilde{g}^{\text{ART}}(\mathbf{x})=\sum_{m=0}^{M}\sum_{r=1}^{R}\left(\sum_{i=n_{0,\text{train}}+1}^{n_0} \frac{w_{m, r, i}}{n_0-{n_{0,\text{train}}}} \right) \hat{g}^{(m, r)}(\mathbf{x}).
\end{equation*}
\end{algorithmic}
\caption{\textbf{ART-I-AM}: ART-Integrated-Aggregating Machines}
\label{alg:ART-I-AM}
\end{algorithm}

ART also provides a natural way of calculating variable importance as long as the algorithm $\mathcal{A}$ outputs a set of variables that are important for prediction. Lasso is utilized to demonstrate the idea in this work, while other sparse penalties like elastic-net \citep{ZH05}, SCAD \citep{FL01}, and MCP \citep{Z10} can be imposed in the algorithm $\mathcal{A}$ to produce sparse coefficients as well. 

With slight abuse of notation, we denote $X_j\in \hat{g}^{(m)}$ if the $j$-th predictor $X_j$ is selected by the model $\hat{g}^{(m)}$ trained by $\mathcal{A}(\tilde{T}^{(m)})$.  For example, variable selection methods (usually enjoy variable selection consistency) like Lasso selects a sparse set of variables. Another example is the random forest, which outputs the variable importance for each feature. In that case, users can determine their own cutoff to select the variables (e.g., top 10 variables or the set of variables with importance greater than 3). Note that if the method $\mathcal{A}$ does not carry out variable selection, the ART variable importance is not well-defined.
\begin{definition}
The \textit{ART variable importance} for the $j$th predictor is defined as 
$$
\mathrm{VI}_j=\sum_{m=0}^{M}w_m I(X_j\in \hat{g}^{(m)} ),
$$
where $I(\cdot)$ is the indicator function and $w_m$ is given in Algorithm~\ref{alg:ART}.
\label{def:VI}
\end{definition}

The ART variable importance cannot lay down a dichotomy rule about whether a predictor is important, while it is a relative importance measure for each predictor's contribution to the final model $\tilde{g}^{\text{ART}}$, bearing a resemblance to the variable importance that arises in random forest and boosting. 

\section{Theory}\label{sec:theory}
In this section, we establish the non-asymptotic statistical learning theory of ART. 

We first introduce some notations. The generalization error of an estimated function $\hat{g}$ is assessed based on a given loss $EL(y, \hat{g}(\mathbf{x}))$, where the expectation $E$ is taken over  $(\mathbf{x},y)$ that is drawn from $(\mathbf{X},Y)$ and all pairs of data that used to generate $\hat{g}$. For example, $\hat{g}^{(1)}$ is trained based on the data $\{(\mathbf{x}^{(0)}_i, y^{(0)}_i)\}$ and $T^{(1)}$, and then the expectation in $EL(y, \hat{g}^{(1)}(\mathbf{x}))$ is taken over all $(\mathbf{x},y)$ that is drawn from $(\mathbf{X},Y)$, and $T^{(0)}_\text{train} \cup T^{(1)}$.
Let $|\cdot |$ denote the Euclidean norm. For a function $g$, let $||g||_2:=\sqrt{E|g(\mathbf{X})|^2}=\sqrt{\int|g(\mathbf{X})|^2P_\mathbf{X}(d\mathbf{X})}$ denote the $L_2$ norm of $g$ with respect to the distribution of $\mathbf{X}$. When no confusion arises, we write the final model of Algorithm~\ref{alg:ART}, $\tilde{g}^{\text{ART}}$, as $\tilde{g}$, for the sake of exposition.

\subsection{ART for regression}
Assume the primary data $T^{(0)}=\{\mathbf{x}^{(0)}_i, y^{(0)}_i\}_{i=1}^{n_0}$ are i.i.d. realizations of the random vector $(\mathbf{X},Y)$, where $\mathbf{X}=(X_1,...,X_p)\in \mathbb{R}^p$ and $Y\in \mathbb{R}$. Denote the conditional mean function as $g(\mathbf{X})=E(Y|\mathbf{X})$ and the conditional variable function as  $\sigma^2(\mathbf{X})=\mathrm{Var}(
Y|\mathbf{X})$. We consider the data-generating model
\begin{equation}\label{model}
Y=g(\mathbf{X})+\sigma(\mathbf{X})\epsilon,
\end{equation} 
where, without loss of generality, the error term $\epsilon \in \mathbb{R}$ has $E(\epsilon|\mathbf{X})=0$ and $\mathrm{Var}(\epsilon|\mathbf{X})=1$.  
\begin{assumption}[Boundedness]\label{ass:31}
Assume that the mean function, the estimated mean functions, and the variance function are upper bounded, i.e., there exists a positive constant $A$ such that $|g(\mathbf{X})|\le A$, $\sup_m|\hat{g}^{(m)}(\mathbf{X})|\le A$, and $\sigma(\mathbf{X})\le A$ almost surely.
\end{assumption}
The above boundedness assumption about mean and variance functions are mild and common in the model averaging/aggregation literature.
\begin{assumption}\label{ass:32}
The loss function $L(a,b)$ is convex in $b$ and can be written in the form $\rho(a-b)$ for some function $\rho(\cdot)$. In addition, there exists two positive constants $c_1$ and $c_2$ such that $2c_{1}|t_{1}-t_{2}|\le|\rho'(t_{1})-\rho'(t_{2})|\le2c_{2}|t_{1}-t_{2}|$ and $c_{1}(t_{1}-t_{2})^{2}\le\rho(t_{1})-\rho(t_{2})-\rho'(t_{2})(t_{1}-t_{2})\le c_{2}(t_{1}-t_{2})^{2}$ for $t_{1},t_{2}\in\mathbb{R}$. 
\end{assumption}
Assumption \ref{ass:32} requires that the function $\rho$ has a Lipschitz continuous first-order derivative, and its second-order derivative is bounded. Many loss functions satisfy this general assumption, for example, the squared error loss $L(a,b)=(a-b)^2$ and the asymmetric error loss $L_\tau(a,b)=|\tau-I(a-b<0)|\cdot (a-b)^2$ with $\tau\in(0,1)$.
\begin{assumption}\label{ass:33}
Given $\mathbf{X}$, the noise $\epsilon$ is sub-exponential. A random variable $Z\in\mathbb{R}$ is called a sub-exponential variable \citep{Vershynin} if its sub-exponential norm is bounded, i.e., $\sup_{k\ge1}k^{-1}(E|Z|^{k})^{1/k}<\infty$.
\end{assumption}
If the noise term $\epsilon$ is independent of $\mathbf{X}$, then the assumption reduces to require a sub-exponential noise. The sub-exponential family contains random variables whose moment generating functions exist in a neighborhood of 0. This is a large and general class.
\begin{theorem}\label{thm:regression}
Recall that $\hat{g}^{(0)}$ is the estimate without transfer learning. Under Assumptions \ref{ass:31}, \ref{ass:32}, and \ref{ass:33}, the excess risk of the final estimate $\tilde{g}$ in Algorithm \ref{alg:ART} is upper bounded as 
\begin{equation}
EL(Y,\tilde{g}^{\text{ART}}(\mathbf{X}))-EL(Y,\hat{g}^{(0)}(\mathbf{X}))\le\frac{\log M}{\lambda\tilde{n}},     
\end{equation}
where $\tilde{n} = n_{0}-{n_{0,\text{train}}}$, if the tuning parameter $\lambda$ satisfies $\lambda^{-1}\ge[16\sqrt{2}d_{1}^{2}\exp(e^{2}d_{1}^{2}/2)+32c_{2}^{2}A^{2}\exp(1/(16e^{2}d_{1}))]\cdot\exp(2\lambda c_{2}A^{2})/2c_{1}\}$ and $\lambda^{-1}\ge8e\cdot d_{1}A$, where $d_{1}=2c_2A\sup_{k\ge1}k^{-1}(E|\epsilon|^{k})^{1/k}$.
\end{theorem}
As indicated by Theorem~\ref{thm:regression}, the excess prediction risk of the proposed estimator $\tilde{g}$ is upper bounded by a small penalty term $1/\{\lambda(n_0-{n_{0,\text{train}}} )\}\cdot \log M$. The penalty term is the price to pay to achieve the best performance without knowing which auxiliary data set is the best.  
Hence, our method still has good performance in the case where all auxiliary data give terrible transferred predictions, i.e., the negative transfer occurs, because the non-transferred model $\hat{g}^{(0)}$ (obtained by training the primary data $T^{(0)}$ without any auxiliary data) is included in the candidate pool. In other words, our method does not require the strong assumption that the auxiliary data should be transferable. Note that the upper bound is generally loose. We can further improve the transferred estimate $\tilde{g}$ in line 2 of Algorithm \ref{alg:ART}. Currently, for each auxiliary data, we simply stack it with the primary data and obtain a transferred estimate $\mathcal{A}(T^{(0)}_{\text{train}})$. Different transfer learning algorithms have their own way of utilizing the auxiliary data. If we know a transfer learning method that has a much smaller excess risk, we can obtain $\mathcal{A}(T^{(0)}_{\text{train}})$ following that transfer learning method. In that case, the excess risk of our final transferred estimate will also improve. We want to emphasize that our algorithm applies to any machine learning method, at the cost of a loose bound in the theorem.

\begin{remark}\label{rem:lam}
The tuning parameter $\lambda$ should not be too small; otherwise, the penalty term will be much larger than $EL(Y,\hat{g}^{(0)}(\mathbf{X}))$ and the upper bound becomes meaningless. In terms of the convergence rate, one only needs to make the penalty term $1/\{\lambda(n_0-{n_{0,\text{train}}} )\}\cdot \log M=O(1/({n_{0,\text{train}}}+n_m))$ since the term $EL(Y,\hat{g}^{(0)}(\mathbf{X}))$ converges to $EL(Y,g(\mathbf{X}))$ at a rate no faster than $1/({n_{0,\text{train}}}+n_m)$. Thus the penalty term will not affect the convergence rate. In practice, we recommend the value $\lambda=({n_{0,\text{train}}}+n_m)/(n_0-{n_{0,\text{train}}})$ when at least one of the auxiliary data sizes is much larger than the primary data size; otherwise, we recommend $\lambda=1$. In our simulations and real-data analysis, the choice $\lambda=1$ works very well. Indeed, the proposed method is stable against the choice of $\lambda$ in a wide range.
\end{remark}

\subsection{ART for classification}
Suppose the primary data $T^{(0)}=\{\mathbf{x}^{(0)}_i, y^{(0)}_i\}_{i=1}^{n_0}$ are i.i.d. realizations of the random vector $(\mathbf{X},Y)$, where $\mathbf{X}=(X_1,...,X_p)\in \mathbb{R}^p$ and $Y\in {0,1}$. Let $g(\mathbf{X})=P(Y=1|\mathbf{X})$ be the conditional probability of $Y$ being 1 given the features $\mathbf{X}$. Let the $m$th trained model $\hat{g}^{(m)}$ be an estimated function of $f$. Then the classifier for each new data point $\textbf{x}_{\text{new}}$ is taken as $I(\hat{g}^{(m)}(\textbf{x}_{\text{new}})>0.5)$. 

\begin{assumption}\label{ass:classi}
 For each auxiliary data $T^{(m)}$, there exists a positive constant $0<A_{m}<0.5$ such that $\hat{g}^{(m)}(\mathbf{x})\in(A_{m},1-A_{m})$ for any $\mathbf{x}$. 
\end{assumption}
The constant sequence $A_{m}$ is allowed to converge to 0 as $m\to \infty$. Assumption \ref{ass:classi} is mild since the case  $g(x)=0$ or $1$ is trivial for classification. 
\begin{theorem}\label{thm:classi}
Under Assumption \ref{ass:classi}, the excess risk of the final estimate $\tilde{g}^{\text{ART}}$  is upper bounded as 
\begin{equation}
E||\tilde{g}^{\text{ART}}-g||_{2}^{2}-\frac{2}{A_{m}^{2}}E||\hat{g}^{(0)}-g||_{2}^{2}\le2\frac{\log M}{n_{0}-{n_{0,\text{train}}}}.
\end{equation}
\end{theorem}

Unlike the regression setting, we present the bound for the squared error loss. This is because when evaluating a classifier $\hat{g}(\mathbf{X})\in {0,1}$, the mean error probability $EP(Y\neq \hat{g}(\mathbf{X}))$ is commonly considered. If the underlying conditional probability $g(\mathbf{X})=P(Y=1|\mathbf{X}=\mathbf{X})$ is known, the mean error probability is minimized by the Bayes classifier $g^*(\mathbf{X}):=I(g(\mathbf{X})>0.5)$. It is then natural to consider the following measure for evaluating $g$: $EP(Y\neq \hat{g}(\mathbf{X}))-P(Y\neq g^*(\mathbf{X}))$. Let $\hat {g}$ be an estimator of $f$. It is known that for any plug-in classifier $I(\hat{g}(\mathbf{x})>0.5)$, it satisfies $EP(Y\neq \hat{g}(\mathbf{X}))-P(Y\neq g^*(\mathbf{X}))\le 2(E||f-\hat{g}||_2^2)^{1/2}$. The plug-in classifier $I(\tilde{g}(\mathbf{X})>0.5)$ in our method also satisfies this inequality.

\subsection{ART variable importance}
Theoretical properties of ART variable importance that is proposed in Definition~\ref{def:VI} are derived in this section.
\begin{theorem}\label{importance}
Define $\mathcal{S}$ to be the unknown set that contains all the important features in the conditional mean function $g(\cdot)$. If there exists some $m\in\{0,1,...,M\}$ such that 
\begin{equation}\label{niceproperty}
    P(X_j \in \mathcal{A}(\hat{g}^{(m)}))\rightarrow 1
\end{equation}as $n\rightarrow \infty$ for any $X_j\in \mathcal{S}$, we have $\min_{j:X_j\in \mathcal{S}} \mathrm{VI}_j\xrightarrow{P} 1$ and $\max_{j:X_j\notin \mathcal{S}} \mathrm{VI}_j\xrightarrow{P} 0.$
\end{theorem}

The ART variable importance depends on $\mathcal{A}$. As Theorem \ref{importance} stated, ART will retain the variable selection property of the original algorithm. The theorem does not hold if $\mathcal{A}$ does not provide variable importance measure (e.g., SVM) or the nice property \eqref{niceproperty} (e.g., random forest). 

\section{Simulation Studies}
\label{sec:simulations}

\subsection{ART for regression}\label{sec:simu_reg}
With $p = 10$ and $n = 50$, we construct the primary data $T^{(0)}$ by independently sampling each $\mathbf{x}_i^{(0)}$ from $\mathrm{N}_p(0, \Sigma)$, where $i = 1, 2, \ldots, n$ and $\Sigma$ possesses the auto-regression correlation structure, i.e., $\Sigma = (0.5^{|i-j|})_{p \times p}$. Generate each response as $y_i^{(0)} = \beta^\top\mathbf{x}_i^{(0)}  + \epsilon$, where $\epsilon$ is from the standard normal distribution and independent of the predictors. The true coefficient of interest, $\beta$, is drawn from $\mathrm{N}_p(\mathbf{1}, \mathbf{I}_p)$. To simulate the auxiliary data $T^{(m)}$, each 
 $\mathbf{x}_i^{(m)}$ is produced in the same way as $\mathbf{x}_i^{(0)}$, and $y_i^{(m)} =\beta^{(m)\top} \mathbf{x}_i^{(m)}  + \epsilon$, where $\beta^{(m)} = \beta + \xi$ and $\xi$ indicates the noise level.

We consider three methods of handling the auxiliary data: (1) fit the least squares regression only on the primary data, giving rise to $\hat{\beta}^{\text{LS}}$, (2) grow a pooled data by stacking the primary data and all the $M$ auxiliary data together, and fit the least squares regression on the pooled data, yielding $\hat{\beta}^{\text{pool-LS}}$, and (3) apply ART to obtain $\hat{\beta}^{\text{ART-LS}}$. To assess the performance, we independently generate another 5,000 observations as test data following the same distribution of the primary data $T^{(0)}$ and evaluate the prediction error $\sum_{i=1}^{5000}(\hat{y}_i - y_i)^2/5000$ according to the test data.

\begin{figure}[t]
\vskip -0.1in
\begin{center}
\centerline{\includegraphics[width=0.6\columnwidth]{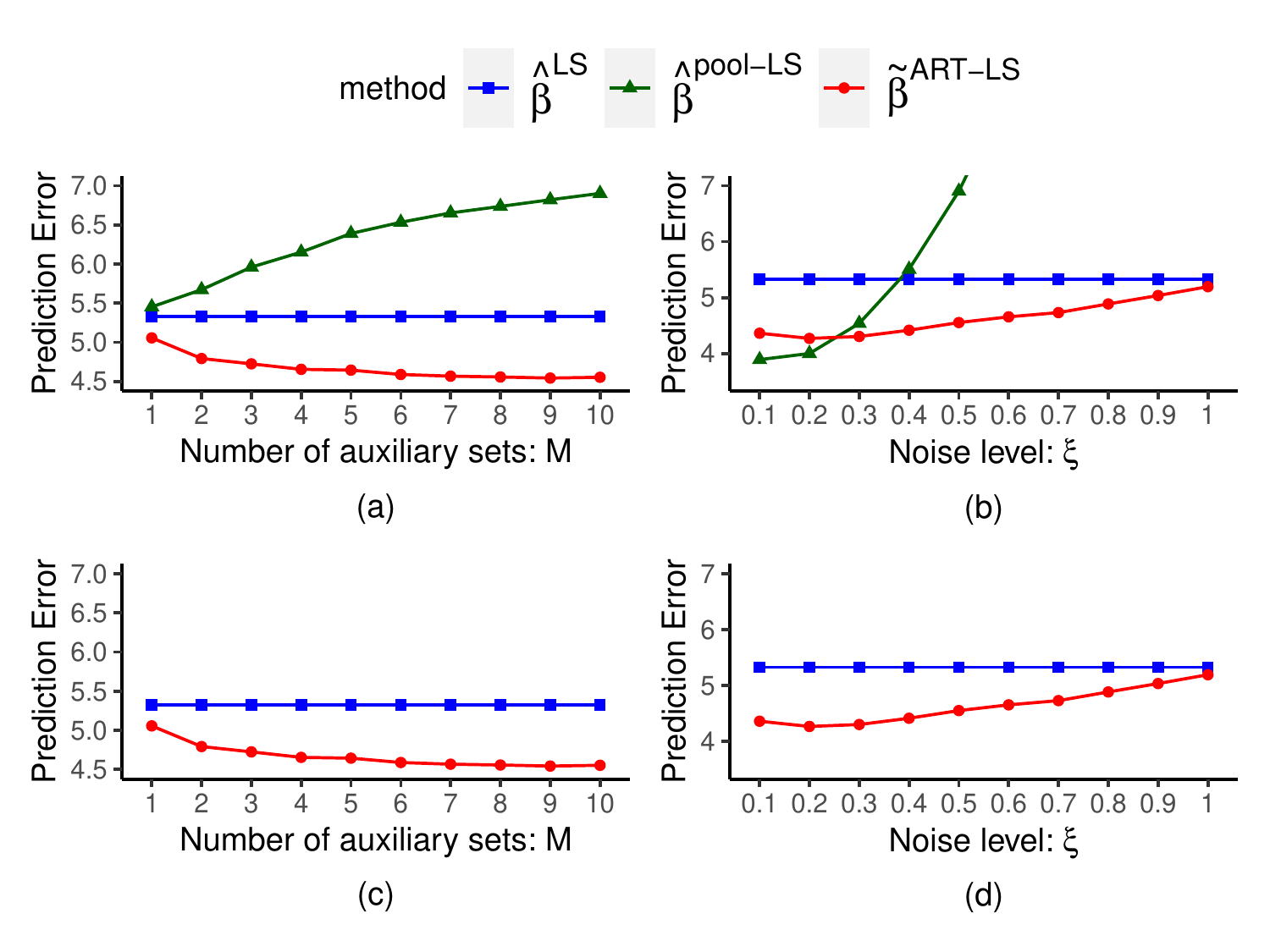}}
\caption{ART for least squares regression. \textit{Example 4.1.1.} Panel (a): prediction errors against $M$, the number of auxiliary data, with the noise level $\xi$ fixed to be $0.5$; panel (b): prediction errors against $\xi$ with $M$ fixed to be $10$. \textit{Example 4.1.2.} Ten adversarial data are included. Panel (c): prediction errors against $M$; panel (d): prediction errors against $\xi$. The prediction errors of $\hat{\beta}^{\text{pool-LS}}$ are not shown because they are far above the errors of the other two estimators. All the errors are averaged over 50 runs with the standard error shown as vertical bars.}
\label{fig:PE_reg}
\end{center}
\vskip -0.2in
\end{figure}

\textbf{Example 4.1.1}\ We study the prediction errors of the three methods by varying $M$ from $1$ to $10$, and the noise level $\xi$ is fixed to be $0.5$. As shown in the panel (a) of Figure~\ref{fig:PE_reg}, the prediction error of $\hat{\beta}^{\text{ART-LS}}$ stays lower than that of $\hat{\beta}^{\text{LS}}$ starting at $M=1$, and declines quickly as $M$ increases. This observation indicates that ART can effectively gain useful information from the auxiliary data. In contrast, the prediction error of $\hat{\beta}^{\text{ART-LS}}$ grows with $M$, hence directly treating the primary and auxiliary data equally brings more noise and exacerbates the accuracy of $\hat{\beta}^{\text{pool-LS}}$. The performance with different noise levels of the auxiliary data is further investigated. With $M=10$ and $\xi$ ranging from $0.1$ to $1$, the panel (b) of Figure~\ref{fig:PE_reg} exhibits the prediction errors: the error curve of $\hat{\beta}^{\text{ART-LS}}$ is quite flat when the noise level $\xi$ is small, indicating that ART consistently improves the original estimator with relatively low-level noise emerging in the auxiliary data; the prediction error approaches, but does not exceed, the error of the $\hat{\beta}^{\text{LS}}$ when $\xi$ rises. Nevertheless, the prediction accuracy of $\hat{\beta}^{\text{pool-LS}}$ drops rapidly as $\xi$ increases. Hence, when the auxiliary data are noisy, ART automatically switches the focus on the primary data and prevents the negative transfer, rather than na\"ively trust the auxiliary data as does $\hat{\beta}^{\text{pool-LS}}$. 

\textbf{Example 4.1.2}\ To further illustrate the robustness of ART against noisy data, we generate ten additional adversarial data, $\check{T}^{(m)}$, which are of the same size as the primary data $T^{(0)}$. For each $m=1,2,\ldots, 10$ and $i=1,2,\ldots, n$, the predictors $\check{\mathbf{x}}_i^{(m)}$ in $\check{T}^{(m)}$ are generated in the same way as the primary data ${T}^{(0)}$. The response is obtained from $\check{y}_i^{(m)} = \check{\beta}^{(m)\top} \check{\mathbf{x}}_i^{(m)}  + \epsilon$ where $\check{\beta}^{(m)} = -{\beta}^{(0)} - \xi$, that is, the coefficients in the auxiliary data $T^{(m)}$ and the adversarial data $\check{T}^{(m)}$ are the opposite. The panel (c) of Figure~\ref{fig:PE_reg} displays the prediction errors against $M$, the number of auxiliary data. Remarkably, the prediction errors of $\hat{\beta}^{\text{ART-LS}}$ are almost identical to the case when the adversarial data are absent. This observation demonstrates the robustness of the ART framework against noisy or even adversarial data. On the other hand, the prediction errors $\hat{\beta}^{\text{pool-LS}}$ are extremely high. Even when $M$ is as high as $10$, the prediction error is $42.26$, and it further rises to $162.41$ when $M = 1$. The pooled estimator $\hat{\beta}^{\text{pool-LS}}$ breaks down when the external data resources are excessively noisy.

\subsection{ART for classification}\label{sec:simu_class}
\textbf{Example 4.2.1}\ We first consider a commonly used linear classifier -- logistic regression. Similar to Section~\ref{sec:simu_reg}, with $p = 10$, we independently generate each predictor $\mathbf{x}_i$ from the same distribution $\mathrm{N}_p(0, \Sigma)$ for one primary data, $M$ auxiliary data, and 10 adversarial data, all of which are of the size $n = 50$. For the primary data, each binary response is drawn from the Bernoulli distribution with the probability $P(y_i = 1) = 1/(1 + \exp(-\pi_i))$, where $\pi_i = \beta^\top \mathbf{x}_i$ and $\beta$ is generated from $\mathrm{N}_p(\mathbf{1}, \mathbf{I}_p)$. The binary responses in the auxiliary and adversarial data are generated in the same way as the primary data, except the coefficients are different: for the $m$th auxiliary data, $\beta^{(m)} = \beta + \xi$, and the $m$th adversarial data, $\check{\beta}^{(m)} = -\beta - \xi$, where $\xi$ is the noise level. We independently generate 5000 data points forming the test data to evaluate the prediction accuracy. 

\begin{figure}[t]
\begin{center}
\centerline{\includegraphics[width=0.6\columnwidth]{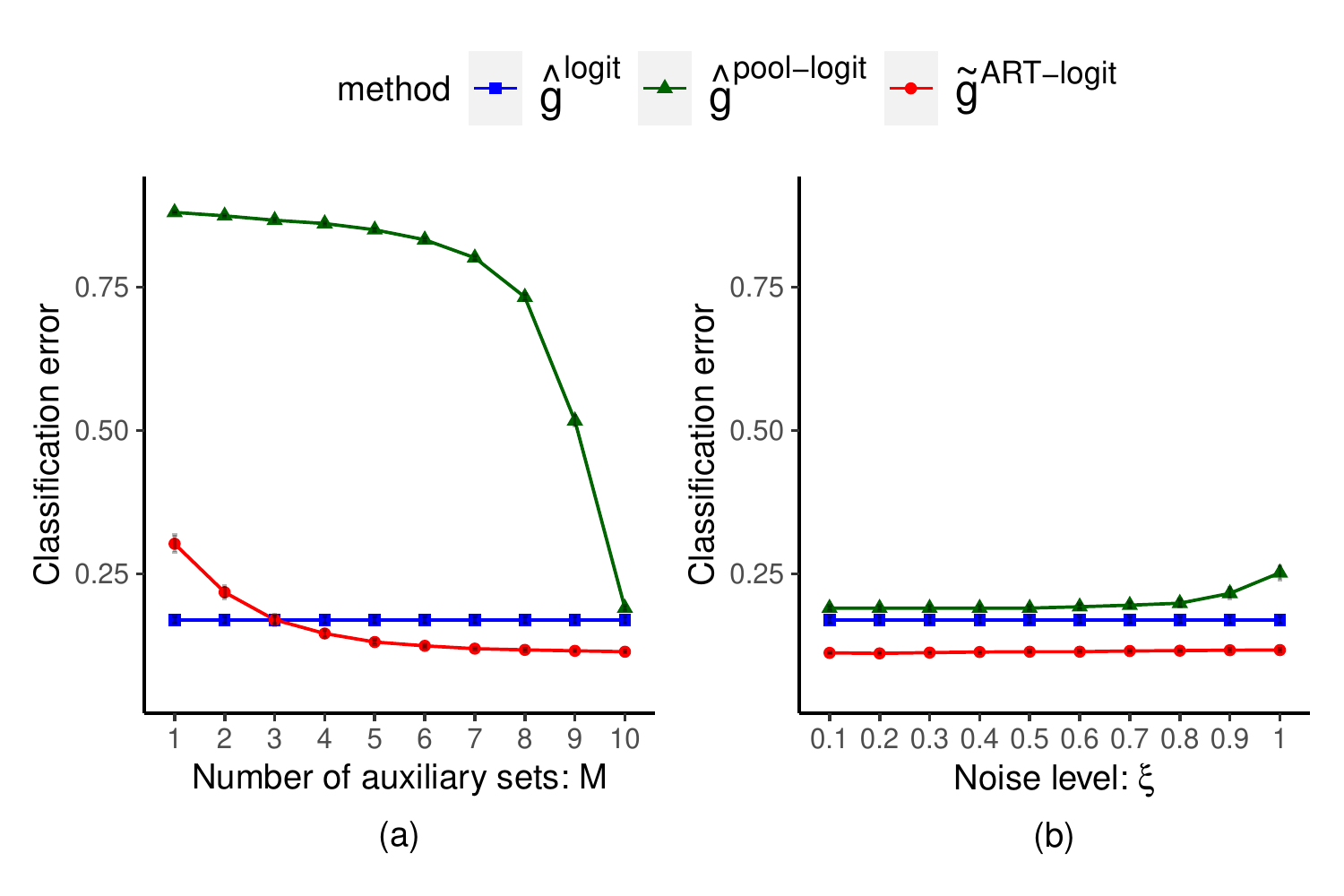}}
\caption{ART for logistic regression. \textit{Example 4.2.1.} Panel (a): classification errors against $M$, the number of auxiliary sets, with the noise level $\xi$ fixed to be $0.5$; panel (b): classification errors against $\xi$ with $M$ fixed to be $10$.}
\label{fig:clas_PE}
\end{center}
\vskip -0.4in
\end{figure}

We fit logistic regression, $\hat{g}^{\text{logit}}$, $\hat{g}^{\text{pool-logit}}$, $\hat{g}^{\text{ART-logit}}$, on the primary data, on the pooled data, and with ART, respectively. The panel (a) of Figure~\ref{fig:clas_PE} plots the prediction error over $M$ with $\xi = 0.5$. We see $\hat{g}^{\text{pool-logit}}$ is greatly affected by the noise in external data resource, and $\hat{g}^{\text{ART-logit}}$ outperforms $\hat{g}^{\text{logit}}$ when $M > 2$. The panel (b) of Figure~\ref{fig:clas_PE} plots the prediction error over $\xi$ with $M = 5$, and it reveals the robustness of ART against the noise level, while $\hat{g}^{\text{pool-logit}}$ incurs higher error as $\xi$ increases.

\textbf{Example 4.2.2}\ To illustrate the wide applicability of ART, we consider more general classifiers. The primary data are generated from a Gaussian mixture distribution. The positive class is assemble by $\sum_{j=1}^5 0.2 \mathrm{N}(\mu_{j+}, \Sigma)$ with each $\mu_{j+}$ drawn from $\mathrm{N}(\mu_+, 1)$, where $\mu_+ = \mu(1, \ldots, 1, 0, \ldots, 0)$, and the negative class is generated according to $\sum_{j=1}^5 0.2 \mathrm{N}(\mu_{j-}, \Sigma)$ with each $\mu_{j-}$ drawn from $\mathrm{N}(\mu_-, 1)$, where $\mu_+ = -\mu(1, \ldots, 1, 0, \ldots, 0)$. We set $\mu = 1$. The auxiliary and adversarial data are generated in the same way, except the auxiliary data have $\mu^{(m)} = \mu + \xi$ while the adversarial data have $\check{\mu}^{(m)} = -\mu - \xi$. We fit random forest, kernel SVM, AdaBoost, and neural nets, using the R packages randomForest \cite{LW02}, magicsvm \citep{WZ22}, gbm \citep{GBC22}, and nnet \citep{nnet}, respectively, with tuning parameters selected based on cross-validation or out-of-bag errors. We see that ART enhances the four classifiers fitted on the primary data as soon as $M > 1$ as shown in the left panel, and it is robust against the adversarial data as observed in the right panel. Among the four classifiers, neural nets perform better than the rest, and ART advances the accuracy on top of them. In addition, ART-I-AM also outperforms the SVM and AdaBoost, and performs similarly to the neural nets, which deliver the best accuracy among the four classifiers.

\subsection{ART sparse learning}\label{sec:simu_hd}
ART can naturally account for high-dimensional data analysis when it is coupled with sparse penalized methods such as lasso. We compare ART with lasso which is fitted with the R package glmnet \citep{FHT10} and the trans-lasso method proposed in \citet{LCL21}. 

\textbf{Example 4.3.1}\ 
Following simulation settings in \citet{LCL21}, we set the dimension $p = 200$, the sample size for the primary and each auxiliary data, $n_0 = 150$ and $n_m = 100$, respectively. All $\mathbf{x}_i$ are drawn from $\mathrm{N}(0, 1)$. To generate the sparse coefficients, for the primary data, the first 16 coordinates of $\beta$ are $0.3$ and all the others are zeros. For the $m$th auxiliary data, $m=1,2,\ldots, M$, define a set $\mathcal{H}^{(m)}$ including the first 16 coordinates and 12 other randomly selected coordinates, and let $\beta^{(m)} = \beta + \mathbf{v}$, where $v_j = 2\xi$ if $j \in \mathcal{H}^{(m)}$ or $0$ otherwise, for each $j = 1, 2, \ldots, 200$. 

\begin{figure}[t]
\begin{center}
\centerline{\includegraphics[width=0.6\columnwidth]{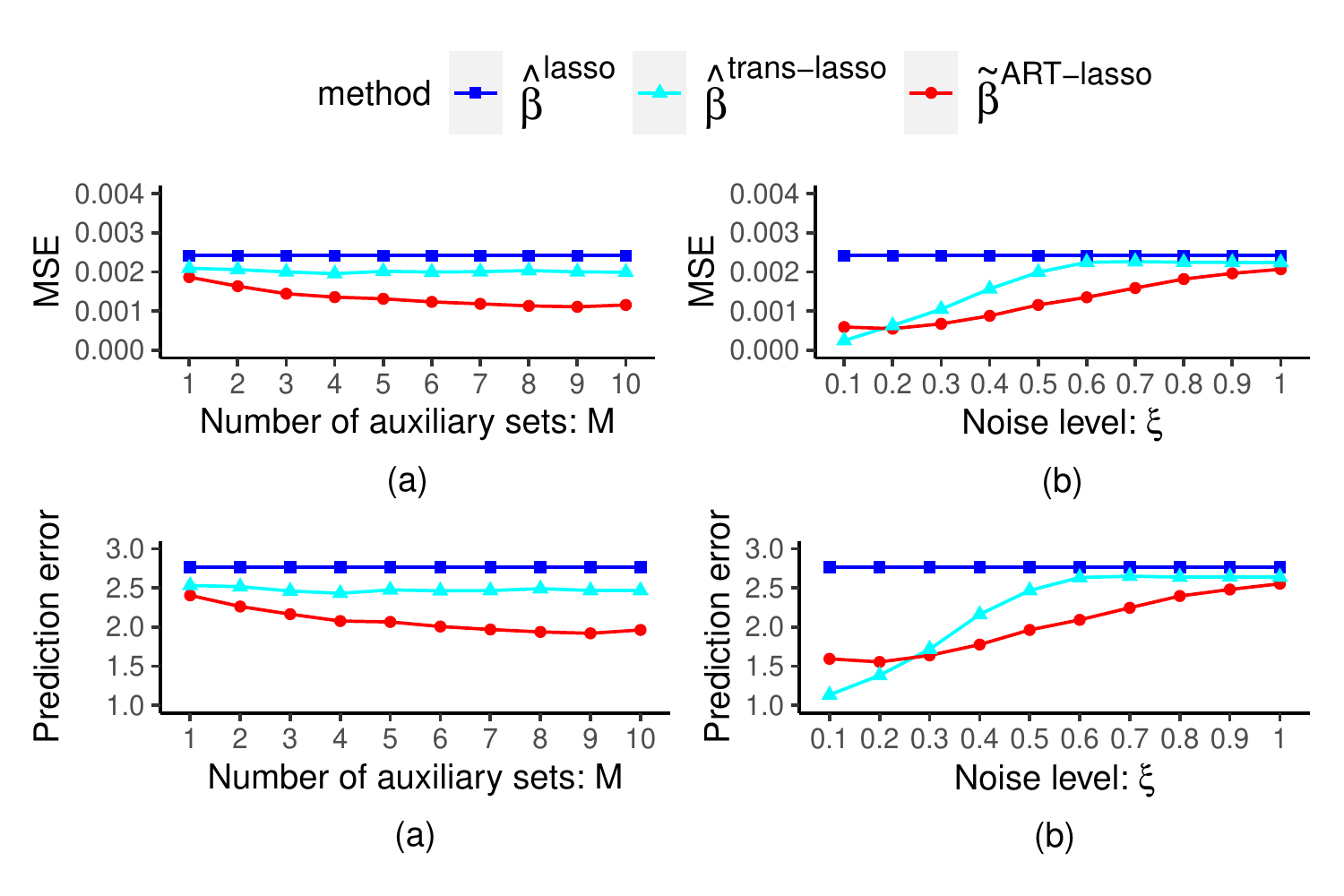}}
\caption{ART for lasso-penalized high-dimensional regression. \textit{Example 4.3.1.} Panel (a): prediction errors against $M$ with $\xi = 0.5$; panel (b): prediction errors against $\xi$ with $M=10$.}
\label{fig:ART_highD}
\end{center}
\vskip -0.2in
\end{figure}

Plotted in the panel (a) of Figure~\ref{fig:ART_highD} are the prediction errors of $\hat{\beta}^{\text{lasso}}$, $\hat{\beta}^{\text{trans-lasso}}$, and $\hat{\beta}^{\text{ART-lasso}}$ over $M$, and plotted in the panel (b) are the prediction errors over the noise level $\xi$. It is observed that $\hat{\beta}^{\text{ART-lasso}}$ consistently improves $\hat{\beta}^{\text{lasso}}$ in all the examples, and $\hat{\beta}^{\text{lasso}}$ outperforms $\hat{\beta}^{\text{trans-lasso}}$ when more auxiliary data are available. ART is more robust over high noise levels than the trans-lasso approach.

\begin{figure}[t]
\begin{center}
\centerline{\includegraphics[width=0.7\columnwidth]{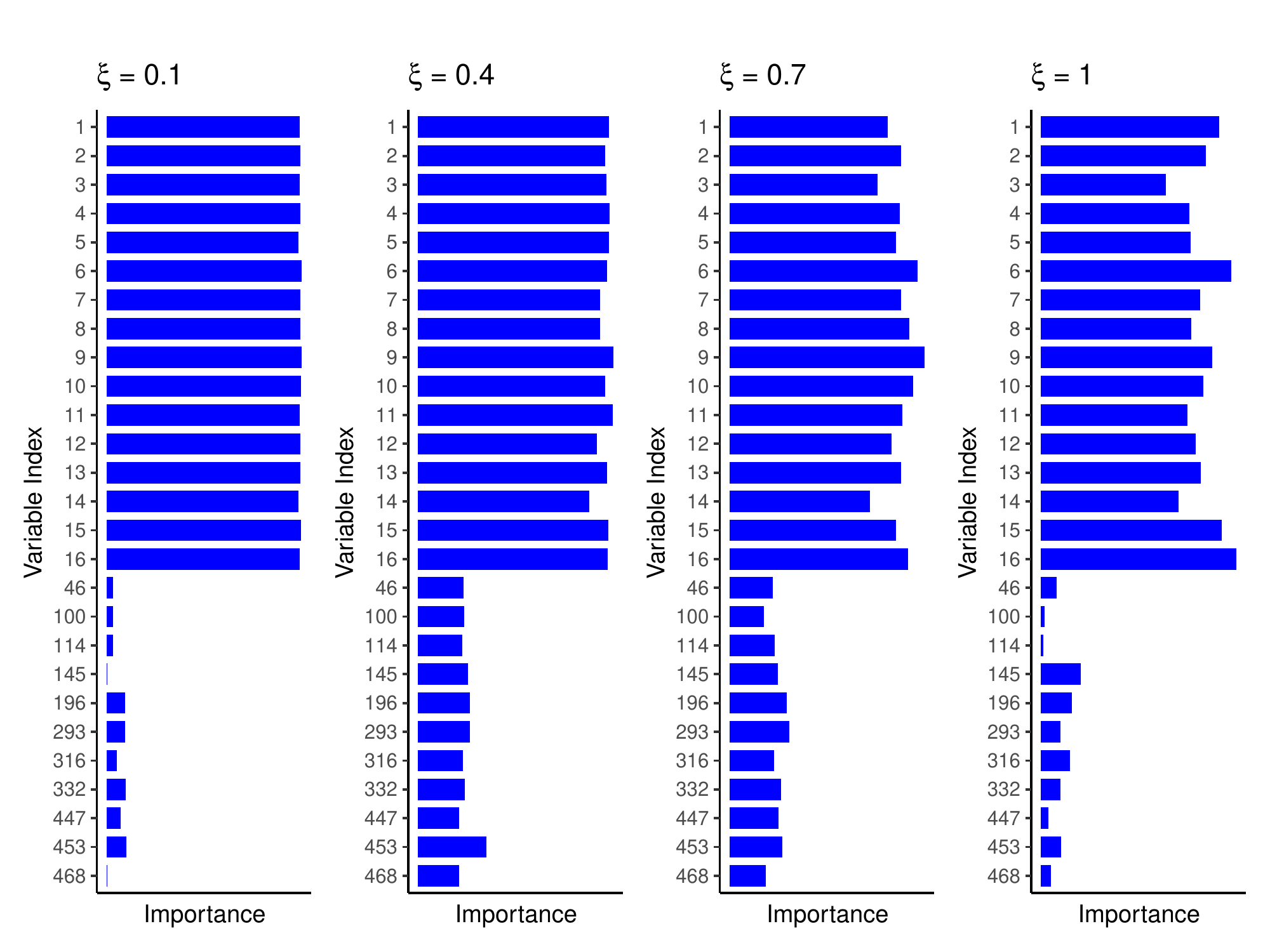}}
\caption{A relative importance spectrum of ART in Example 4.3.1, with $M = 5$ and $\xi$ varies from $0.1, 0.4, 0.7$, and $1$. Only the first 16 variables are designed to be active. The variables not shown have zero importance in all the examples. }
\label{fig:ART_VI}
\end{center}
\vskip -0.2in
\end{figure}

With $M$ fixed to be $5$, Figure~\ref{fig:ART_VI} depicts a relative importance spectrum obtained from ART, where the noise level $\xi=0.1, 0.4, 0.7, 1$ are used to exemplify the results. ART sets a clear cut-off in the importance of the first 16 active coefficients, and the growing noise level brings only little increase in the relative importance of the inactive coefficients.

\section{A Real Application on Intensive Care Unit Mortality}
\label{sec:real_data}
Intensive care units (ICU) are lifesaving for urgent and life-threatening patients, thanks to advanced therapeutic and monitoring technologies, as well as large provider-to-patient ratios. On the other hand, the lifesaver comes with a high cost. Since back in 2005, the total cost of ICU has been around 30\% of the healthcare budget and been over 0.66\% of the gross domestic product \citep{HP10, HP15}, making ICU a highly limited healthcare resource. It is of critical importance to triage patients with a reasonable estimate of the post-ICU survival rate. The outcome can help patients and their families better prepare the possible long-term ICU care that may have poor outcomes despite high expenses, or help providers decide on an alternative of ICU, e.g., progressive care \citep{stacy2011progressive}.

In this work, we apply ART to study the mortality rate of the ICU. We collected data from a multi-center database, eICU Collaborative Research Database \citep{pollard2018eicu}, comprising de-identified health data across the United States between 2014 and 2015. To predict the survival status of the individuals, we only used their information when they were at the ICU admission and excluded all the discharge information. We grouped the patient ages into six categories: infants \& children (0--12), teens (13--19), early adults (20--39), middle-aged adults (40--59), senior adults (60--89), and extremely senior adults ($>$89). To reduce the admission diagnosis categories, we merged the categories with low mortality rates as "others". 

Our goal is to predict the post-ICU survival status of the teaching hospitals in the data. By grouping the hospitals by their region (Midwest, Northeast, South, and West) and capacity (Large if over 500 beds and Small otherwise) and using the undersampling method to have the same number of survived and decreased patients, we generated four sub-data, Midwest-Small ($n=72$, mortality rate $p=9.47\%$), Northeast-Large ($n=1230$, $p=8.84\%$), South-Large ($n=644$, $p=6.93\%$), South-Small ($n=210$, $p=4.29\%$), and West-Large ($n=372$, $p=6.59\%$). The estimation of the survival status of the Midwest-Small sub-data is of particular interest due to its relatively high mortality rate, while the problem is challenging because of the relatively small sample size. Consequently, we treat Midwest-Small as the primary data and appoint the other three sub-data as the auxiliary data in the ART framework. To assess the performance, we randomly split the primary data by having 50 patients for training and the rest for evaluating the classification error. ART is applied on random forest, AdaBoost, and neural nets, while SVM is not included due to its high computational cost. With 50 random splits, the averaged classification error is exhibited in Table~\ref{realdata}, where we see AdaBoost delivers lower classification error than random forest and neural nets when these classifiers are fitted on only the primary data. The errors of random forest and neural nets decrease when the classifiers are fitted on the pooled data stacking the primary and the three auxiliary data, whereas the pooling strategy increases the error of AdaBoost. This observation reflects the potential risk of the negative transfer. For all three classifiers, ART outperforms both the original and pooled algorithms. Last, we observe that ART-I-AM delivers the same classification accuracy as the best classifier. 

\section{Conclusion}
In this work, we propose ART, an adaptive and robust pipeline designed for applying transfer learning to general statistical and machine learning methods. Through examples of regression, classification, and sparse learning methods, we demonstrate that ART can effectively enhance estimation by extracting and digesting useful information from auxiliary data, while remaining resilient to noisy data sources. 

As ART is a general pipeline that is not specifically tailored to any particular methods, it would be intriguing to explore its performance on large-scale image and text data with deep learning algorithms in future research. 

\clearpage

\vspace{-5cm}

\appendix

\section{Techinical Proofs}
\subsection{Proof of Theorem \ref{thm:regression}.}
Denote 
\begin{equation}
q_{{n_{0,\text{train}}}}^{n_{0}}=\sum_{m=0}^{M}\pi_{m}\exp\left\{ -\lambda\sum_{i={n_{0,\text{train}}}+1}^{n_{0}}L\left(y_{i},\hat{g}^{(m)}(\mathbf{x}_{i}^{(0)})\right)\right\} .\label{eq:qm}
\end{equation}
We can decompose $q_{{n_{0,\text{train}}}}^{n_{0}}$ as

\begin{align}
q_{{n_{0,\text{train}}}}^{n_{0}}\nonumber  =&\sum_{m=0}^{M}\pi_{m}\exp\left\{ -\lambda L\left(y_{{n_{0,\text{train}}}},\hat{g}^{(m)}(\mathbf{x}_{{n_{0,\text{train}}}+1}^{(0)})\right)\right\} \nonumber  \times\frac{\sum_{m=0}^{M}\pi_{m}\exp\left\{ -\lambda\sum_{i={n_{0,\text{train}}}+1}^{{n_{0,\text{train}}}+2}L\left(y_{i},\hat{g}^{(m)}(\mathbf{x}_{i}^{(0)})\right)\right\} }{\sum_{m=0}^{M}\pi_{m}\exp\left\{ -\lambda L\left(y_{{n_{0,\text{train}}}+1},\hat{g}^{(m)}(\mathbf{x}_{{n_{0,\text{train}}}+1}^{(0)})\right)\right\} }\nonumber \\
 & \times\cdots\times\frac{\sum_{m=0}^{M}\pi_{m}\exp\left\{ -\lambda\sum_{i={n_{0,\text{train}}}+1}^{n_{0}}L\left(y_{i},\hat{g}^{(m)}(\mathbf{x}_{i}^{(0)})\right)\right\} }{\sum_{m=0}^{M}\pi_{m}\exp\left\{ -\lambda\sum_{i={n_{0,\text{train}}}+1}^{n_{0}-1}L\left(y_{i},\hat{g}^{(m)}(\mathbf{x}_{i}^{(0)})\right)\right\} }\nonumber \\
= & \prod_{i={n_{0,\text{train}}}+1}^{n_{0}}\left(\sum_{m=0}^{M}w_{m,i}\exp\left\{ -\lambda L\left(y_{i},\hat{g}^{(m)}(\mathbf{x}_{i}^{(0)})\right)\right\} \right).\label{eq:qn_decomp}
\end{align}
Let $J$ be a discrete random variable with $P(J=m)=w_{m,i},m\ge0$,
where $i\in\{{n_{0,\text{train}}}+1,...,n_{0}\}$ is fixed. Let $\nu$ be the discrete
measure induced by $J$ on $\mathbb{Z}^{+}$ such that $\nu(m)=P(J=m)$.
Denote $h(J)=-L(y_{i}^{(0)},\hat{g}^{(J)}(\mathbf{x}_{i}^{(0)}))$. We have that 
$$
\sum_{m=0}^{M}w_{m,i}\exp\left\{ -\lambda L\left(y_{i}^{(0)},\hat{g}^{(m)}(\mathbf{x}_{i}^{(0)})\right)\right\} =E_{\nu}\exp(\lambda h(J)).
$$
By Lemma 3.6.1 of \citet{30555}, we have 
$$
\log E_{\nu}\exp(\lambda h(J))\le\lambda E_{\nu}h(J)+\frac{\lambda^{2}}{2}\mathrm{Var}_{\nu}(h(J))\exp\left(\lambda\max\left\{0,\sup_{\gamma\in[0,\lambda]}\dfrac{M_{\nu_{\gamma}}^{3}(h(J))}{\mathrm{Var}_{\nu_{\gamma}}(h(J))}\right\}\right),
$$
where the discrete measure 
$$
\nu_{\gamma}(m)=\frac{w_{m,i}\exp(\gamma h(m))}{\sum_{m=1}^{M}w_{m,i}\exp(\gamma h(m))}
$$
for $m\ge0$ and 
$
M_{\nu_{\gamma}}^{3}(h(J))=E_{\nu_{\gamma}}(h(J)-E_{\nu_{\gamma}}(h(J)))^{3}.
$
Then
\begin{align*}
  \sup_{\gamma\in[0,\lambda]}\frac{M_{\nu_{\gamma}}^{3}(h(J))}{\mathrm{Var}_{\nu_{\gamma}}(h(J))}
\le & \sup_{\gamma\in[0,\lambda]}\sup_{m\ge0}|h(m)-E_{\nu_{\gamma}}h(J)|\\
\le & 2\sup_{m\ge0}\left|L\left(y_{i}^{(0)},\hat{g}^{(m)}(\mathbf{x}_{i}^{(0)})\right)-L\left(y_{i}^{(0)},g(\mathbf{x}_{i}^{(0)})\right)\right|\\
\le & 2|\rho'(y_{i}^{(0)}-g(\mathbf{x}_{i}^{(0)}))|\cdot\sup_{m\ge0}|\hat{g}^{(m)}(\mathbf{x}_{i}^{(0)})-g(\mathbf{x}_{i}^{(0)})|+2c_{2}\sup_{m\ge0}(\hat{g}^{(m)}(\mathbf{x}_{i}^{(0)})-g(\mathbf{x}_{i}^{(0)}))^{2}\\
\le & 2|\rho'(\sigma(\mathbf{x}_{i}^{(0)})\epsilon_{i})|\cdot A+2c_{2}A^{2}
\end{align*}
and 
\begin{align*}
\mathrm{Var}_{\nu}(h(J)) & \le E_{\nu}\left(L\left(y_{i}^{(0)},\hat{g}^{(J)}(\mathbf{x}_{i}^{(0)})\right)-L\left(y_{i}^{(0)},E_{\nu}\hat{g}^{(J)}(\mathbf{x}_{i}^{(0)})\right)\right)^{2}\\
 & \le\sup_{j\ge0}\left(\rho'(y_{i}^{(0)},\hat{g}^{(j)}(\mathbf{x}_{i}^{(0)})) +c_{2}|\hat{g}^{(j)}(\mathbf{x}_{i}^{(0)})-E_{\nu}\hat{g}^{(J)}(\mathbf{x}_{i}^{(0)})|\right)^{2}E_{v}\left(\hat{g}^{(J)}(\mathbf{x}_{i}^{(0)})-E_{\nu}\hat{g}^{(J)}(\mathbf{x}_{i}^{(0)})\right)^{2}\\
 & \le\sup_{j\ge0}\left(\rho'(y_{i}^{(0)},g(\mathbf{x}_{i}^{(0)}))+4c_{2}\sup_{j\ge0}|\hat{g}^{(j)}(\mathbf{x}_{i}^{(0)})-g(\mathbf{x}_{i}^{(0)})|\right)^{2} \cdot E_{v}\left(\hat{g}^{(J)}(\mathbf{x}_{i}^{(0)})-E_{\nu}\hat{g}^{(J)}(\mathbf{x}_{i}^{(0)})\right)^{2}\\
 & \le \left(\rho'(\sigma(\mathbf{x}_{i}^{(0)})\epsilon_{i})+4c_{2}A\right)^{2}E_{v}\left(\hat{g}^{(J)}(\mathbf{x}_{i}^{(0)})-E_{\nu}\hat{g}^{(J)}(\mathbf{x}_{i}^{(0)})\right)^{2},
\end{align*}
where the first two inequalities hold by Assumption \ref{ass:33}. Furthermore, by Assumption \ref{ass:33},
\begin{align*}
 & E_{v}\left[L\left(y_{i}^{(0)},\hat{g}^{(J)}(\mathbf{x}_{i}^{(0)})\right)-L\left(y_{i}^{(0)},E_{\nu}\hat{g}^{(J)}(\mathbf{x}_{i}^{(0)})\right)\right]\\
\ge& E_{v}\left[\rho'\left(y_{i}^{(0)}-E_{\nu}\hat{g}^{(J)}(\mathbf{x}_{i}^{(0)})\right)\left(E_{\nu}\hat{g}^{(J)}(\mathbf{x}_{i}^{(0)})-\hat{g}^{(J)}(\mathbf{x}_{i}^{(0)})\right)\right] +c_{1}E_{v}\left(E_{\nu}\hat{g}^{(J)}(\mathbf{x}_{i}^{(0)})-\hat{g}^{(J)}(\mathbf{x}_{i}^{(0)})\right)^{2}\\
=&c_{1}E_{v}\left(E_{\nu}\hat{g}^{(J)}(\mathbf{x}_{i}^{(0)})-\hat{g}^{(J)}(\mathbf{x}_{i}^{(0)})\right)^{2}
\end{align*}
which leads to 
$$
E_{v}(E_{\nu}\hat{g}^{(J)}(\mathbf{x}_{i}^{(0)})-\hat{g}^{(J)}(\mathbf{x}_{i}^{(0)}))^{2}\le\frac{1}{c_{1}}E_{v}\left[L\left(y_{i}^{(0)},\hat{g}^{(J)}(\mathbf{x}_{i}^{(0)})\right)-L\left(y_{i}^{(0)},E_{\nu}\hat{g}^{(J)}(\mathbf{x}_{i}^{(0)})\right)\right].
$$ 
We have 
\begin{align}\label{eq:above}
 \log E_{v}\exp\{\lambda h(J)\}
\le & -\lambda E_{\nu}L(y_{i}^{(0)},\hat{g}^{(J)}(\mathbf{x}_{i}^{(0)}))+\frac{\lambda^{2}}{2}(\rho'(\sigma(\mathbf{x}_{i}^{(0)})\epsilon_{i})^{2}+16c_{2}^{2}A^{2})\cdot\exp\left\{ 2\lambda|\rho'(\sigma(\mathbf{x}_{i}^{(0)})\epsilon_{i})|\cdot A\right\} \nonumber\\
 & \cdot\frac{1}{c_{1}}E_{v}\left[L\left(y_{i}^{(0)},\hat{g}^{(J)}(\mathbf{x}_{i}^{(0)})\right)-L\left(y_{i},E_{\nu}\hat{g}^{(J)}(\mathbf{x}_{i}^{(0)})\right)\right]\cdot\exp\{2\lambda c_{2}A^{2}\}.
\end{align}

Since any sub-exponential variable $Z$ satisfies 
$
E\exp\left(t|Z|\right)\le2\exp(2e^{2}d_{1}^{2}t^{2})
$ and 
$
E\left[Z^{2}\exp\left\{ t|Z|\right\} \right]\le d_{4}\exp(d_{3}t^{2})
$ 
for any $|t|\le d_2/d_1$, where $d_{1}=\sup_{k\ge1}k^{-1}(E|Z|^{k})^{1/k}$, $d_{2}=1/(4e)$, $d_{3}=8e^{4}d_{1}^{2}$, $d_{4}=16\sqrt{2}d_{1}^{2}$. Here, redefine 
$$
d_{1}=\sup_{k\ge1}k^{-1}(E_{\epsilon|\mathbf{X}}|\rho'(\sigma(\textrm{\textbf{X}}))\cdot\epsilon|^{k})^{1/k}\le 2c_2A\sup_{k\ge1}k^{-1}(E|\epsilon|^{k})^{1/k}.
$$
By Assumption \ref{ass:33}, the above inequalities hold for $Z:=\epsilon$ given $\mathbf{X}$.
Taking expectation $E_{y_{i}|\mathbf{x}_{i}^{(0)},(\mathbf{x}_{k}^{(0)},y_{k}^{(0)})_{k=1}^{i-1},T^{(J)}}$ (denoted as $E_{i}$ for convenience) of both sides of the 
inequality \eqref{eq:above}, for $2\lambda A\le d_2/d_1$ we have 
\begin{align*}
 & E_{i}\log E_{v}\exp\{\lambda h(J)\}\\
\le & -\lambda E_{i}E_{\nu}L(y_{i}^{(0)},\hat{g}^{(J)}(\mathbf{x}_{i}^{(0)}))+\frac{\lambda^{2}}{2}E_{i}(d_{2}\exp(d_{3}(2\lambda A)^{2})+32c_{2}^{2}A^{2}\exp(d_{1}(2\lambda A\text{)}^{2}))\\
 & \cdot\frac{1}{c_{1}}E_{v}\left[L\left(y_{i}^{(0)},\hat{g}^{(J)}(\mathbf{x}_{i}^{(0)})\right)-L\left(y_{i}^{(0)},E_{\nu}\hat{g}^{(J)}(\mathbf{x}_{i}^{(0)})\right)\right]\cdot\exp\{2\lambda c_{2}A^{2}\}.
\end{align*}
By choosing a small enough $\lambda$ such that 
$$
\frac{\lambda^{2}}{2}\frac{\exp\{2\lambda c_{2}A^{2}\}}{c_{1}}E_{i}(d_{2}\exp(d_{3}(2\lambda A)^{2})+32c_{2}^{2}A^{2}\exp(d_{1}(2\lambda A)^{2}))\text{\ensuremath{\le\lambda}},
$$
which gives
\begin{align*}
 E_{i}\log E_{v}e^{\lambda h(J)}
\le  -\lambda E_{i}E_{\nu}L(y_{i}^{(0)},\hat{g}^{(J)}(\mathbf{x}_{i}^{(0)})) +\lambda E_{i}E_{v}\left[L\left(y_{i}^{(0)},\hat{g}^{(J)}(\mathbf{x}_{i}^{(0)})\right)-L\left(y_{i}^{(0)},E_{\nu}\hat{g}^{(J)}(\mathbf{x}_{i}^{(0)})\right)\right]
=  -\lambda E_{i}\left[L\left(y_{i}^{(0)},E_{\nu}\hat{g}^{(J)}(\mathbf{x}_{i}^{(0)})\right)\right].
\end{align*}
We have 
\begin{align*}
  E\log(1/q_{{n_{0,\text{train}}}}^{n_{0}})
= & -\sum_{i={n_{0,\text{train}}}+1}^{n_{0}}E\log\left(\sum_{m=0}^{M}w_{m,i}\exp\left\{ -\lambda L\left(y_{i}^{(0)},\hat{g}^{(m)}(\mathbf{x}_{i}^{(0)})\right)\right\} \right)\\
= & -\sum_{i={n_{0,\text{train}}}+1}^{n_{0}}EE_{i}\log\left(E_{v}\exp\left\{ -\lambda L\left(y_{i}^{(0)},\hat{g}^{(J)}(\mathbf{x}_{i}^{(0)})\right)\right\} \right)\\
\ge & \lambda\sum_{i={n_{0,\text{train}}}+1}^{n_{0}}EL\left(y_{i}^{(0)},E_{v}\hat{g}^{(J)}(\mathbf{x}_{i}^{(0)})\right).
\end{align*}
We also have for each $m\ge0$, 
\begin{align*}
 E\log(1/q_{{n_{0,\text{train}}}}^{n_{0}})
\le & \log(1/\pi_{m})+\lambda\sum_{i={n_{0,\text{train}}}+1}^{n_{0}}EL\left(y_{i}^{(0)},\hat{g}^{(m)}(\mathbf{x}_{i}^{(0)})\right)
=  \log(1/\pi_{m})+\lambda(n_{0}-{n_{0,\text{train}}})EL\left(Y,\hat{g}^{(m)}(\mathbf{X})\right).
\end{align*}
Thus, by convexity, we have 
\begin{align*}
  EL\left(y_{i}^{(0)},\tilde{g}(\mathbf{x}_{i}^{(0)})\right)
\le & \frac{1}{(n_{0}-{n_{0,\text{train}}})}\sum_{i={n_{0,\text{train}}}+1}^{n_{0}}EL\left(y_{i}^{(0)},E_{v}\hat{g}^{(J)}(\mathbf{x}_{i}^{(0)})\right) \frac{\log(1/\pi_{m})}{\lambda(n_{0}-{n_{0,\text{train}}})}+EL\left(Y,\hat{g}^{(m)}(\mathbf{X})\right),
\end{align*}
where the desired result follows.

\subsection{Proof of Theorem \ref{thm:classi}.}
Denote $K_{g}(\mathbf{x},y)=g(\mathbf{x})^{y}(1-g(\mathbf{x}))^{1-y}$ as the joint density function
of $(\mathbf{X},Y)$ with respect to the product measure $\mu\times\nu$, where
$\nu$ is the counting measure on $\{0,1\}$. Denote $\tilde{g}_{i}(\mathbf{x})=\sum_{m}w_{m,i}\hat{g}^{(m)}(\mathbf{x})$,
then
$
K_{\tilde{g}_{i}}(\mathbf{x},y)=\sum_{m}w_{m,i}K_{\hat{g}^{(m)}}(\mathbf{x},y).
$

Take $\lambda=1$ and the loss function in (\ref{eq:qm}) to be $$
L\left(y_{i}^{(0)},\hat{g}^{(m)}(\mathbf{x}_{i}^{(0)})\right):=-y_{i}^{(0)}\log(\hat{g}^{(m)}(\mathbf{x}_{i}^{(0)}))-(1-y_{i}^{(0)})\log(1-\hat{g}^{(m)}(\mathbf{x}_{i}^{(0)})),
$$
we then have $q_{{n_{0,\text{train}}}}^{n_{0}}=\prod_{i={n_{0,\text{train}}}+1}^{n_{0}}K_{\tilde{g}_{i}}(\mathbf{x}_{i}^{(0)},y_{i})$.
Hence, 
\begin{align*}
 & \sum_{i={n_{0,\text{train}}}+1}^{n_{0}}E\int K_{g}(\mathbf{x},y)\log\frac{K_{g}(\mathbf{x},y)}{K_{\tilde{g}_{i}}(\mathbf{x},y)}\mu\times\nu(d\mathbf{x}dy)\\
= & \sum_{i={n_{0,\text{train}}}+1}^{n_{0}}E\int K_{g}(\mathbf{x}_{i}^{(0)},y_{i}^{(0)})\log\frac{K_{g}(\mathbf{x}_{i}^{(0)},y_{i}^{(0)})}{K_{\tilde{g}_{i}}(\mathbf{x}_{i}^{(0)},y_{i}^{(0)})}\mu\times\nu(d\mathbf{x}_{i}^{(0)} dy_{i}^{(0)})\\
= & \sum_{i={n_{0,\text{train}}}+1}^{n_{0}}E\int[\Pi_{i={n_{0,\text{train}}}+1}^{n_{0}}K_{g}(\mathbf{x}_{i}^{(0)},y_{i}^{(0)})]\cdot\log\frac{K_{g}(\mathbf{x}_{i}^{(0)},y_{i}^{(0)})}{K_{\tilde{g}_{i}}(\mathbf{x}_{i}^{(0)},y_{i}^{(0)})}
\mu\times\nu(d\mathbf{x}_{{n_{0,\text{train}}}+1}^{(0)}dy_{{n_{0,\text{train}}}+1}^{(0)})\cdots\mu\times\nu(d\mathbf{x}_{n_{0}}^{(0)}dy_{n_{0}}^{(0)})\\
= & E\int[\Pi_{i={n_{0,\text{train}}}+1}^{n_{0}}K_{g}(\mathbf{x}_{i}^{(0)},y_{i}^{(0)})]\cdot\log\frac{\Pi_{i={n_{0,\text{train}}}+1}^{n_{0}}K_{g}(\mathbf{x}_{i}^{(0)},y_{i}^{(0)})}{q_{{n_{0,\text{train}}}}^{n_{0}}}
\mu\times\nu(d\mathbf{x}_{{n_{0,\text{train}}}+1}^{(0)}dy_{{n_{0,\text{train}}}+1}^{(0)})\cdots\mu\times\nu(d\mathbf{x}_{n_{0}}^{(0)}dy_{n_{0}}^{(0)})\\
\le & E\int[\Pi_{i={n_{0,\text{train}}}+1}^{n_{0}}K_{g}(\mathbf{x}_{i}^{(0)},y_{i}^{(0)})] \cdot\log\frac{\Pi_{i={n_{0,\text{train}}}+1}^{n_{0}}K_{g}(\mathbf{x}_{i}^{(0)},y_{i}^{(0)})}{\pi_{m}\exp\left\{ -\sum_{i={n_{0,\text{train}}}+1}^{n_{0}}L_{0}(y_{i}^{(0)},\hat{g}^{(m)}(\mathbf{x}_{i}^{(0)}))\right\} }
\mu\times\nu(d\mathbf{x}_{{n_{0,\text{train}}}+1}^{(0)}dy_{{n_{0,\text{train}}}+1}^{(0)})\cdots\mu\times\nu(d\mathbf{x}_{n_{0}}^{(0)}dy_{n_{0}}^{(0)})\\
\le & \log(1/\pi_{m})+E\int[\Pi_{i={n_{0,\text{train}}}+1}^{n_{0}}K_{g}(\mathbf{x}_{i}^{(0)},y_{i}^{(0)})] \cdot\log\frac{\Pi_{i={n_{0,\text{train}}}+1}^{n_{0}}K_{g}(\mathbf{x}_{i}^{(0)},y_{i}^{(0)})}{\exp\left\{ -\sum_{i={n_{0,\text{train}}}+1}^{n_{0}}L_{0}(y_{i}^{(0)},\hat{g}^{(m)}(\mathbf{x}_{i}^{(0)}))\right\} }\\
 & \mu\times\nu(d\mathbf{x}_{{n_{0,\text{train}}}+1}^{(0)}dy_{{n_{0,\text{train}}}+1}^{(0)})\cdots\mu\times\nu(d\mathbf{x}_{n_{0}}^{(0)}dy_{n_{0}}^{(0)})\\
= & \log(1/\pi_{m})+\sum_{i={n_{0,\text{train}}}+1}^{n_{0}}E\int K_{g}(\mathbf{x},y)\log\frac{K_{g}(\mathbf{x},y)}{K_{\hat{g}^{(m)}}(\mathbf{x},y)}\mu\times\nu(d\mathbf{x}dy),
\end{align*}
where we use change of variable for the first equality, the fact that $K_{g}$ is a probability density function for the second equality, and the fact that the logarithm is an increasing function in the first inequality. We have 
\begin{align*}
 \int K_{g}(\mathbf{x},y)\log\frac{K_{g}(\mathbf{x},y)}{K_{\hat{g}^{(m)}}(\mathbf{x},y)}\mu\times\nu(d\mathbf{x}dy)
=  \int\left[g(\mathbf{x})\log\frac{g(\mathbf{x})}{\hat{g}^{(m)}(\mathbf{x})}+(1-g(\mathbf{x}))\log\frac{1-g(\mathbf{x})}{1-\hat{g}^{(m)}(\mathbf{x})}\right]\mu(d\mathbf{x})
\le & \frac{1}{A_{m}^{2}}||g-\hat{g}^{(m)}||_{2}^{2},
\end{align*}
where the first inequality follows from the fact that the K-L divergence
is upper bounded by Chi-squared distance ($D_{\mathrm{KL}}(g_{1}||g_{2}):=\int g_{1}(x)\log(g_{1}(x)/g_{2}(x))dx\le\int(g_{1}-g_{2})^{2}/g_{2}(x)dx$
for any probability densities $g_{1}$ and $g_{2}$), and the second
inequality is due to the boundedness assumption. Next, we denote the
squared Hellinger distance between $g_{1}$ and $g_{2}$ as $d_{\mathrm{H}}^{2}(g_{1},g_{2}):=\int(\sqrt{g_{1}}-\sqrt{g_{2}})^{2}dx$.
Then we have 
\begin{align*}
 \frac{1}{2}\sum_{i={n_{0,\text{train}}}+1}^{n_{0}}E||g-\tilde{g}_{i}||_{2}^{2}
\le  \sum_{i={n_{0,\text{train}}}+1}^{n_{0}}Ed_{\mathrm{H}}^{2}(K_{g},K_{\tilde{g}_{i}})
\le  \sum_{i={n_{0,\text{train}}}+1}^{n_{0}}ED_{\mathrm{KL}}(K_{g}||K_{\tilde{g}_{i}})
\le  \log(1/\pi_{m})+\frac{1}{A_{m}^{2}}\sum_{i={n_{0,\text{train}}}+1}^{n_{0}}E||g-\hat{g}^{(m)}||_{2}^{2},
\end{align*}
where the first inequality holds due to
\begin{align}\label{eq:constant}
 Ed_{\mathrm{H}}^{2}(K_{g_{1}},K_{g_{2}})\nonumber
= \int(\sqrt{g_{1}}-\sqrt{g_{2}})^{2}+(\sqrt{1-g_{1}}-\sqrt{1-g_{2}})^{2}\mu(d\mathbf{x})\nonumber
\ge  \frac{1}{4}\int(g_{1}-g_{2})^{2}+(1-g_{1}-(1-g_{2}))^{2}\mu(d\mathbf{x})\nonumber
=  \frac{1}{2}\int(g_{1}-g_{2})^{2}\mu(d\mathbf{x})
\end{align}
for any two bounded functions $0\le g_{1},g_{2}\le1$, and the second inequality holds because the K-L divergence is lower bounded by the squared Hellinger distance. Thus, by convexity of the squared error loss, we have 
\begin{align*}
 E||g-\tilde{g}||_{2}^{2}
\le  \frac{1}{n_{0}-{n_{0,\text{train}}}}\sum_{i={n_{0,\text{train}}}+1}^{n_{0}}E||g-\tilde{g}_{i}||_{2}^{2}
\le  2\inf_{m}\left(\frac{\log(1/\pi_{m})}{n_{0}-{n_{0,\text{train}}}}+\frac{1}{A_{m}^{2}}E||g -\hat{g}^{(m)}||_{2}^{2}\right).
\end{align*}

\subsection{Proof of Theorem \ref{importance}}
Suppose that $m_{0}$ is such that $P(X_{j}\in\hat{g}^{(m_{0})})\rightarrow1$ as $n\rightarrow\infty$, we have \begin{align*}
    E\sum_{j:X_{j}\in \mathcal{S}}\mathrm{VI}_{j}	= 	E\sum_{j:X_{j}\in \mathcal{S}}\sum_{m=0}^{M}w_{m}I(X_{j}\in\hat{g}^{(m)})
	=	\sum_{m=0}^{M}w_{m}\sum_{j:X_{j}\in \mathcal{S}}P(X_{j}\in\hat{g}^{(m)}).
\end{align*}
By \citet{10.1214/009053607000000514}, when the data splitting ratio ${n_{0,\text{train}}}/n_{0}$ is chosen properly (such as ${n_{0,\text{train}}}/n_{0}=1/2$ in our case), the exponential-type weighting $w_{m}$ is consistent in the sense that $w_{m_{0}}\rightarrow1$ and $w_{m}\rightarrow0$ for $m\neq m_{0}$. Hence we have $E\sum_{j:X_{j}\in \mathcal{S}}\mathrm{VI}_{j}\rightarrow |\mathcal{S}|$. Since $0\le \mathrm{VI}_j \le 1$ for any $j$, we have $\min_{j:X_j\in \mathcal{S}} \mathrm{VI}_j{\displaystyle \xrightarrow{P}}1$. The proof of $\max_{j:X_{j}\notin \mathcal{S}}\mathrm{VI}_{j}{\displaystyle \xrightarrow{P}}0$ is similar.

\nocite{*}
\bibliography{ART}

\clearpage

\begin{figure}[t]
\begin{center}
\centerline{\includegraphics[width=0.6\columnwidth]{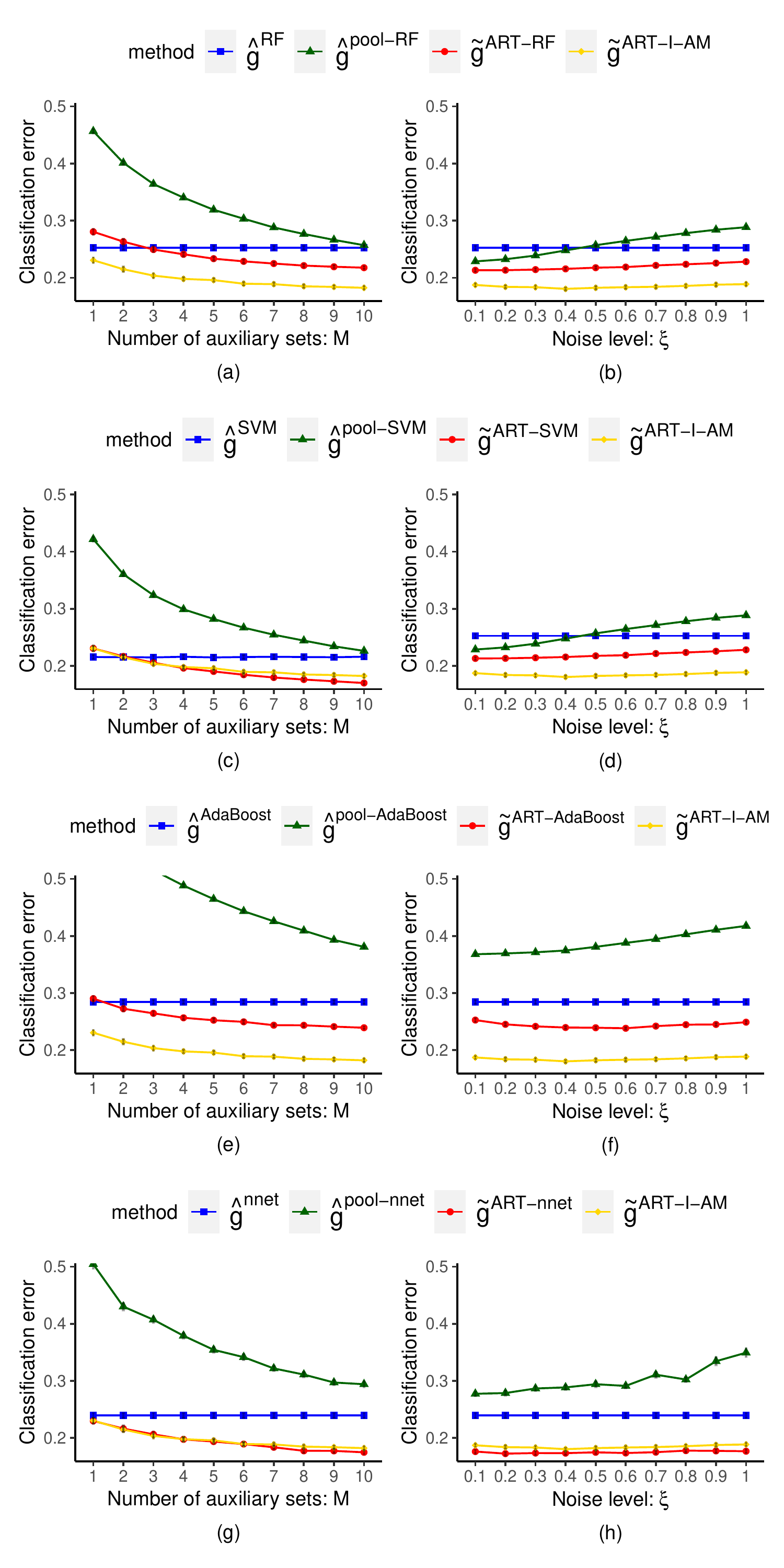}}
\caption{ART for machine learning algorithms. \textit{Example 4.2.2.} Classification errors against $K$ with $\xi = 0.5$ or against $\xi$ with $M=10$ for random forest in panel (a) and (b), for kernel SVM in panel (c) and (d), for AdaBoost in panel (e) and (f), and for neural nets in panel (g) and (h).
}
\label{fig:ART_clas_all}
\end{center}
\vskip -0.2in
\end{figure}

\clearpage

\begin{table}[t]
\caption{Classification error (in percentage) for the real data analysis. Compared are the three transfer learning methods, fitting classifiers on the primary data $\hat{g}^{\text{primary}}$, on the pooled data, $\hat{g}^{\text{pool}}$, and using the ART framework, $\tilde{g}^{\text{ART}}$, applied on random forest, AdaBoost, and neural nets. The lowest classification errors are boldfaced. All the errors are averaged from 50 random splits of the primary data and the standard errors are given.}
\label{realdata}
\vskip -0.1in
\begin{center}
\begin{tabular}{lcccr}
\toprule
method & $\hat{g}^{\text{primary}}$ & $\hat{g}^{\text{pool}}$ & $\tilde{g}^{\text{ART}}$ \\
\midrule
RF    & 44.00$\pm$ 0.01   & 41.36$\pm$ 0.01 & 39.91$\pm$ 0.01  \\
AdaBoost & 41.55$\pm$ 0.01 & 41.73$\pm$ 0.01 & \textbf{38.55}$\pm$ 0.01  \\
nnet    & 48.72$\pm$ 0.01 & 40.73$\pm$ 0.01 & 38.64$\pm$ 0.01  \\
{ART-I-AM}    & -- & -- &       \textbf{38.55}$\pm$ 0.01  \\
\bottomrule
\end{tabular}
\end{center}
\end{table}

\clearpage

\end{document}